# Bridging the Gap: Generalising State-of-the-Art U-Net Models to Sub-Saharan African Populations

Alyssa R. Amod[*,1,2] [0000-0001-5190-9597], Alexandra Smith[3], Pearly Joubert[4] [0000-0001-5735-7455], Confidence Raymond[5,6] [0000-0003-3927-9697], Dong Zhang[7] [0000-0002-2948-1384], Udunna C. Anazodo[5,6,8,9] [0000-0001-8864-035X], Dodzi Motchon[4] [0000-0002-6222-3483], Tinashe E.M. Mutsvangwa[4] [0000-0003-1210-2832], & Sébastien Quetin[10,11] [0009-0005-5391-7319]

[1]Brain Behaviour Unit, Neuroscience Institute, University of Cape Town, Cape Town, ZAF
[2]ShockLab, Dept. of Mathematics & Applied Mathematics, UCT, CT, ZAF
[3]Applied Mathematics Division, Stellenbosch University, Stellenbosch, ZAF
[4]Division of Biomedical Engineering, Neuroscience Institute, UCT, CT, ZAF
[5]Medical Artificial Intelligence Lab, Lagos, NGA
[6]Lawson Health Research Institute, London, Ontario, CAN
[7]Dept. of Electrical & Computer Engineering, University of British Columbia, CAN
[8]Montreal Neurological Institute, McGill University, Montréal, CAN
[9]Dept. of Medicine and Dept. Clinical & Radiation Oncology, University of Cape Town, Cape Town, ZAF
[10]Medical Physics Unit, Dept. of Oncology, McGill University, Montréal, CAN
[11]Montreal Institute for Learning Algorithms, Montréal, CAN

**Abstract.** A critical challenge for tumour segmentation models is the ability to adapt to diverse clinical settings, particularly when applied to poor quality neuroimaging data. The uncertainty surrounding this adaptation stems from the lack of representative datasets, leaving top-performing models without expo-sure to common artefacts found in MRI data throughout Sub-Saharan Africa (SSA). We replicated a framework that secured the 2nd position in the 2022 BraTS competition to investigate the impact of dataset composition on mod-el performance and pursued four distinct approaches through training a model with: BraTS-Africa data only (*train_SSA*, N=60), 2) BraTS-Adult Glioma data only (*train_GLI*, N=1251), 3) both datasets together (*train_ALL*, N=1311), and 4) through further training the *train_GLI* model with BraTS-Africa data (*train_ftSSA*). Notably, training on a smaller low-quality dataset alone (*train_SSA*) yielded subpar results, and training on a larger high-quality dataset alone (*train_GLI*) struggled to delineate oedematous tissue in the low-quality validation set. The most promising approach (*train_ftSSA*) involved pre-training a model on high-quality neuroimages and then fine-tuning it on the smaller low-quality dataset. This approach outperformed the others, ranking 2nd in the MICCAI BraTS Africa global challenge external testing phase. These findings underscore the significance of larger sample sizes and broad exposure to data in improving segmentation performance. Furthermore, we demonstrated there is potential for improving such models by fine-tuning them with a wider range of data locally.

**Keywords:** Sub-Saharan Africa, Brain Tumour Segmentation, MRI, Deep Learning, U-Net.

# 1  Introduction

The burden of cancer continues to rise each year, disproportionately affecting low and middle-income countries (LMICs) due to limited access to imaging technologies as well as specialists, which are crucial for early diagnosis and successful treatment [1,2]. Accurate tumour segmentation provides essential information for treatment decisions, thereby impacting patient survival rate [3,4]. Among intracranial tumours, gliomas are highly heterogeneous, and their prognosis remains poor despite considerable progress in the field of neuro-oncology [5–7]. Gliomas have varied shapes, sizes, boundaries, intensity distributions, and volume fluctuations, which all pose a challenge for accurate segmentation using magnetic resonance imaging (MRI) [8]. MRI is the standard of care and preferred modality for brain tumor imaging [9].

Despite its advantages, various challenges still need to be addressed in low-resource settings, specifically, within sub-Saharan Africa (SSA), such as limited access to high-field MRI scanners and a shortage of qualified experts to acquire, analyze, and interpret MRI data, which leads to delayed tumour diagnosis [5,10]. This coupled with the relatively poor healthcare systems, socio-economic status, and the usual late presentation of disease in SSA, worsens prognosis and sustain the high mortality rates in SSA [1]. These factors continue to drive the growing demand for access to early diagnosis and timely interventions for intracranial tumours. The development of accurate automatic brain tumour segmentation models that can per-form well across varying clinical settings therefore holds significant importance on a global scale.

The field of brain tumour segmentation has evolved considerably in recent decades, aiming to overcome limitations such as reliance on domain expertise, subjectivity, or sensitivity to noise and intensity non-homogeneity [12,13]. Machine learning techniques, specifically deep learning, are at the forefront of this domain, demonstrating significant improvement in terms of segmentation precision between the tumorous tissue, surrounding oedema and normal tissue [14–16]. A combination of convolutional neural networks (CNNs) and autoencoders is currently the best approach and has gradually reduced the dependence on domain expertise and complex feature extraction methods [17]. Specifically, the U-Net model has established itself as a gold-standard in automated brain tumour segmentation tasks [18,19]. Several adaptations of the U-Net have been proposed to further improve efficiency, accuracy, and generalisability [20–23]. However, large neuroimaging datasets required for the generalisation of deep learning models are limited in the medical field, due to both ethical concerns, and data privacy regulations [24].

Since its establishment, the Brain Tumour Segmentation (BraTS) Challenge[1] has witnessed a remarkable surge in globally contributed brain patient data derived from multi-parametric (mp-)MRI scans; comprising the largest annotated publicly available brain tumour dataset in 2023, with ~4,500 cases [25]. The community-developed benchmark for automated brain tumour segmentation fostered through this collabo-ration has significantly contributed to the development of novel architectures and refined training

---

[1] Initiated 2012, in collaboration with the Medical Image Computing and Computer Assisted Interventions (MICCAI) Society

procedures within the domain [3,14,23,26]. However, despite promising performance, it is still uncertain whether these top performing methods can be effectively applied to data obtained from SSA populations, where the quality of MR images remain poor [11,29]. This is largely due to a lack of representative datasets compounded by vast differences in image acquisition parameters across settings, which create challenges for generalisation [17]. Current benchmarks are primarily based on high-resolution brain MRI obtained in standard resource-rich clinical set-tings in high-income countries, incorporating image pre-processing steps that may not expose models to artefacts commonly seen in routine clinical scans from SSA [30].

A key framework currently at the forefront is the nn-UNet developed by Isensee and colleagues [27], which provides a robust tumour segmentation pipeline capable of adapting to various imaging modalities and anatomical structures [28]. This framework demonstrates improved efficacy through ensembling predictions from different U-Net based architectures, with emphasis on the importance of modelling decisions such as pre-processing, data augmentation pipelines, and hyper-parameter tuning approaches. Recently, Zeineldin and colleagues [31] demonstrated that an expectation-maximisation ensemble of the DeepSeg, the nn-UNet, and the Deep-SCAN U-Net based pipelines performs well on a small external test dataset from SSA; achieving Dice score coefficients (DSC) of 0.9737, 0.9593, and 0.9022, for the whole tumour (WT), tumour core (TC) and enhancing tumour (ET), respectively, and Hausdorff distance (95%) (HD95) scores below 3.32 for all sub-regions. However, testing on an external dataset with similar characteristics to the training data did not perform as well, and the difference between validation results for their ensemble approach and the independent run of a nn-UNet based model was minuscule (~0.005, U=6.0, p=0.7). This may further emphasise that a simpler framework with more aggressive data augmentations and carefully selected post-processing methods are likely to play a significant role in developing models that achieve similar performance across a variety of datasets and that can be applied readily in low-resource settings. The former is particularly important for increasing relevant data representations during model training [32].

This year for the first time, the BraTS challenge was expanded to include multimodal MRI training data from low-resource settings. In this work, we examined the extent to which established U-Net tumour segmentation frameworks can train with and generalize to MRI data from low-resource settings to their lower resolution and data quality, and the feasibility of implementing such frameworks with limited re-sources. Given the challenge's goal of creating a versatile benchmark model, we prioritised establishing comparable metrics with top performing methods as a reference point rather than introducing complex augmentations. Our strategy was to replicate a BraTS Challenge benchmark model [37] to investigate the influence of training data composition on the segmentation predictions of an external dataset comprised solely of low-resolution data.

## 2 Methods

### 2.1 Data Description

The dataset used in this study were derived from the BraTS 2023 Challenge data. The training, validation and testing datasets comprised of mp-MRI scans from a total 1565 pre-operative adult glioma patients with 1470 from pre-exiting BraTS data (BraTS-Adult Glioma) as described in previous works [3,25,26,30], and an additional 95 cases from SSA (BraTS-Africa) [11]. These are typical clinical scans obtained from various institutions as part of standard care, leading to significantly diverse imaging quality. Each case included a T1-weighted (T1), post-gadolinium contrast T1-weighted (T1Gd), T2-weighted (T2) and T2 Fluid Attenuated Inversion Recovery (T2-FLAIR). Figure 1 illustrates sample slices obtained from volumes of a patient from each dataset, emphasising the substantial variations in data quality between BraTS-Adult Glioma and BraTS-Africa. The BraTS-Africa dataset was collected through a collaborative network of imaging centres in Africa, with support from the Consortium for Advancement of MRI Education and Research in Africa (CAMERA)[2] and funding from the Lacuna Fund in Health Equity[3]. In each case, image-based ground truth annotations of the tumour sub-regions were generated and approved through an iterative process [11]. Refinement occurred through manual review by volunteers, 2 expert radiologists, and final approval for release was provided by an expert neuro-radiologist. These tumour sub-regions are radiological features and do not reflect strict biological entities [11,25] and include enhancing tumour (ET; label=3), peritumoral oedematous tissue (ED; label=2), and the necrotic core (NCR; label=1), while the voxels not annotated are considered as background (label=0).

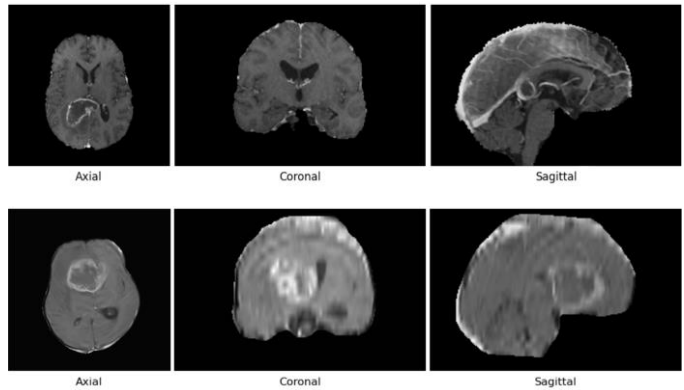

**Fig. 1.** Axial, coronal, and sagittal slices of T1Gd MRIs from the high-resolution BraTS-Adult Glioma dataset (*top*) and the low-resolution BraTS-Africa dataset (*bottom*).

---

[2] https://www.cameramriafrica.org/
[3] https://lacunafund.org

## 2.2 Selecting a Framework

Since the development of the nn-UNet pipeline [27], it has undergone several revisions which primarily focused on the training parameters rather than the architecture itself. Table 1 (rows 1-3) show challenge results for the original nn-UNet model from 2017, and revised models submitted by Isensee et al. in subsequent BraTS challenges, where it placed 2nd in 2018 [33] and 1st in 2020 [34]. The nn-UNet effectively addresses the challenges of manually accounting for co-dependencies when making design changes and selecting the best performing ensemble (see further [28]). How-ever, some teams participating in the BraTS challenge in 2021 [35, 36] and 2022 [31, 37] demonstrated that the basic 3D U-Net component of nn-UNet could be optimised to achieve good performance on brain tumour segmentation tasks, without the need to run the full ensemble of networks encompassed in the nn-UNet pipeline. A comparison of results (see Table 1, rows 4-7 showing external validation results) produced by these teams primarily shows that there is minor difference in overall model performance when sample size remains the same. In comparison to the varied performance seen in the three nn-UNet submissions trained with different smaller sample sizes, this may further highlight the importance of the sample size used for training segmentation models and its effect on model performance.

**Table 1.** BraTS Challenge results: final test scores for nn-UNet trained with different samples, and validation set scores for nn-UNet based models trained with the same sample[a].

| Year [ref.][b] | $N$ | Model | Ave DSC (*HD95*) | WT DSC (*HD95*) | TC DSC (*HD95*) | ET DSC (*HD95*) |
|---|---|---|---|---|---|---|
| 2017 [27] | 285 | nn-UNet | 82.1 *(-)* | 85.8 *(-)* | 77.5 *(-)* | 64.7 *(-)* |
| 2018 [33] | 285[c] | nn-UNet | 84.33 *(5.64)* | 87.81 *(6.03)* | 80.62 *(5.08)* | 77.88 *(2.90)* |
| 2020 [34] | 369 | nn-UNet | 85.35 *(14.55)* | 88.95 *(8.50)* | 85.06 *(17.34)* | 82.03 *(17.81)* |
| 2021 [35] | 1251 | 3D UNet | 91.56 *(1.72)* | 94.86 *(1.41)* | 94.25 *(1.73)* | 90.81 *(2.00)* |
| 2021 [36] | 1251 | 3D UNet + attention | 88.36 *(10.61)* | 92.75 *(3.47)* | 87.81 *(7.62)* | 84.51 *(20.73)* |
| 2022 [31] | 1251 | Ensemble | 88.21 *(9.54)* | 92.71 *(3.60)* | 87.53 *(7.53)* | 84.38 *(17.50)* |
| 2022 [37] | 1251 | 3D UNet | 88.25 *(7.96)* | 92.92 *(3.59)* | 88.02 *(5.84)* | 83.81 *(14.46)* |

[a]Results obtained from published papers, except [35], which is obtained from the Synapse online evaluation platform for 2021; [b]Year of BraTS challenge and citation. [c]Final model submitted was co-trained with external dataset ($N = 484$).

Our selection took into account both simplicity and efficacy, as easy replication in resource-limited environments would be more beneficial for translation to local settings in the long term. In addition to sample size, Futrega et al. [35] demonstrated that alterations to network architecture make little difference in terms of overall tumour segmentation performance. Specifically, they show that varying network layers (e.g., residual connections, multi-head self-attention) or integrating different architectures (e.g., a residual U-Net with an autoencoder or a vision transformer with basic U-Net) remain comparable with a basic 3D U-Net architecture (mean Dice score across sub-regions ranging ~0.002) [35]. This is further supported by a comparison of the ensemble models in [31] and [36], which show no significant difference between the nn-UNet component and their ensemble results ($p>0.7$). We therefore chose to implement the framework outlined by Futrega and colleagues from their 2022 BraTS challenge participation and we refer the reader to both their original paper [35], and their recently released paper [37], which details the full pipeline of the framework.

## 2.3 Data Pre-Processing

All data were provided after initial pre-processing by the challenge organisers to ensure the removal of all protected health information prior to the public sharing of data. The BraTS standardised pipeline detailed in [3, 25] was used, and pre-processing steps included conversion of the DICOM files to NifTI format, to strip all personal patient metadata from headers; skull stripping, to deface neuroimaging scans; co-registration to the SRI24 anatomical template; and finally, re-sampling to a uniform 1mm3 isotropic resolution (see also [11]). We then implemented several additional pre-processing steps as outlined by the OptiNet pipeline [35]. These steps included: stacking the volumes from all modalities; cropping the redundant back-ground voxels, to reduce computational cost; normalisation of the non-zero regions; and, adding a foreground one hot encoding channel, to differentiate between tumorous and non-tumorous regions. This resulted in a final input tensor of shape (5,240,240,155) with channels representing the 4 modalities and the one-hot encoding layer.

## 2.4 OptiNet Pipeline and Experiments

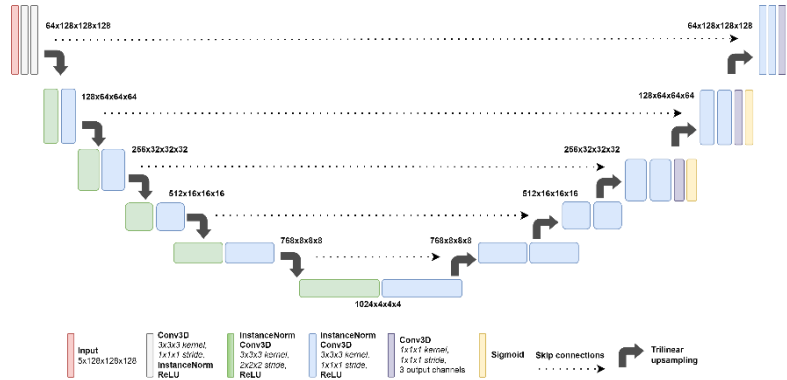

**Fig. 2.** Tuned OptiNet model architecture replicated from [37].

OptiNet primarily adjusts the nn-UNet framework in terms of modelling choices related to pre-processing of data, data augmentations implemented, loss function applied (region-based, summing binary cross-entropy and Dice loss), and several post-processing steps. Data augmentations applied included both spatial transforms (random crop of size 128x128x128, random flip) and intensity related transforms (Gaussian noise, Gaussian blur, changes in brightness) and were implemented during training. Yousef et al. [38] provides a detailed review of the basic U-Net architecture: briefly, it consists of a contracting (encoder) and expansive (decoder) path that are used for down sampling and up sampling, respectively. In the OptiNet version, the encoder, made up of a standard CNN, partners with a decoder that combines feature maps from convolutional layers and trilinear up-sampling. Cropped volumes of size 128x128x128 with preserved context undergo skip connections to reduce information loss between the encoder and decoder, aiding the decoder in restoring image resolution and spatial structure. Figure 2 depicts the model architecture used in our experiments, reproduced from the open access notebooks provided[4]. Futrega et al. [37] include minor architectural changes (e.g., channel sizes, up sampling technique) to the default UNet architecture from nn-UNet. A full review of all architectural and non-architectural tweaks explored by Futrega and colleagues are detailed their 2021 [35] and 2022 [37] papers. Notably, their results showed only a slight increase in mean Dice score when combining some of these changes with the additional one-hot encoding channel (~0.0026, $p$=0.5; see Table 3 in [35]).

We ran four experiments using the OptiNet pipeline by varying the composition of the training dataset. We trained these models with 1) BraTS-Africa data only (*train_SSA*, N=60), 2) BraTS-Adult Glioma data only (*train_GLI*, N=1251), 3) both datasets together (*train_ALL*, N=1311), and 4) through further training the *train_GLI* model with BraTS-Africa data (*train_ftSSA*). The *train_ftSSA* model was included to obtain an estimate of the feasibility of exporting pre-trained benchmark models for fine-tuning locally in Africa. All trainings and validations were conducted with Pytorch 1.9 on a high-performance computing cluster provided by Compute Canada, using one GPU of either NVIDIA T4 Turing (16GB GDDR6 memory) or NVIDIA V100 Volta (16GB/32GB HBM2 memory), depending on the availability on the cluster. All trainings were performed with an internal validation procedure, using a split of the training dataset. Due to time and computational constraints, we trained models containing the larger BraTS-Adult Glioma dataset for 100 epochs on one-fold each. For models containing only the smaller BraTS-Africa dataset (*train_SSA* and *train_ftSSA*), training was set for a maximum of 150 epochs per fold with a 5-fold cross validation and an early stopping strategy when there was no improvement in DSC on internal validation in 100 epochs. Table 2 shows mean dice scores achieved during training per fold. Checkpoints giving best dice scores on the internal validation set of each fold were used for external validation predictions for each experiment.

**Table 2.** Average Dice scores from 5-fold cross validation for OptiNet models trained with BraTS-Adult Glioma data and our models trained with BraTS-Africa data.

| 5-fold CV | OptiNet 2021 | OptiNet 2022 | train _SSA | train _ALL | train _GLI | train _ftSSA |
|---|---|---|---|---|---|---|
| Fold 0 | 91.18 | 91.48 | 93.48 | 90.64 | 90.18 | 89.83 |
| Fold 1 | 91.41 | 91.5 | 88.64 | - | - | 92.62 |
| Fold 2 | 91.76 | 92.17 | 89.35 | - | - | 95.04 |
| Fold 3 | 92.68 | 92.74 | 89.62 | - | - | 93.64 |

---

[4] https://github.com/NVIDIA/DeepLearningExamples/tree/master/PyTorch/Segmentation/nnUNet

| | | | | | | |
|---|---|---|---|---|---|---|
| Fold 4 | 90.76 | 91.05 | 82.15 | - | - | 87.30 |
| **Mean Dice** | **91.56** | **91.79** | **88.65** | **90.64** | **90.18** | **91.69** |

## 3 Results

Below we describe the performance metrics obtained for each of the 4 training variations, with validation scores computed through the challenge online evaluation system. Metrics provided by the system (see Table 3) are lesion-wise dice score coefficients (DSC) and Hausdorff distance (95%) (HD95) averaged across the 15 patient samples provided for the BraTS-Africa 2023 validation phase. No ground truth segmentations are provided; however, the small size of the dataset allowed for a more detailed review of the generated segmentation masks for all validation subjects. The figures presented in this section show the segmentations we generated from our models on selected validation cases. All segmentations are represented by the non-biological labels provided for annotation as described earlier, where ED is green, ET is blue, NCR is red.

Table 3. Segmentation performance on BraTS-Africa validation set (*N*=15)[a].

| **Model** | **DSC** | | | | **HD95** | | | |
|---|---|---|---|---|---|---|---|---|
| | Average | WT | TC | ET | Average | WT | TC | ET |
| *train_SSA* | 59.11 (*31.65*) | 61.34 (*31.52*) | 58.94 (*31.70*) | 57.05 (*31.74*) | 130.15 (*126.56*) | 126.31 (*128.30*) | 126.34 (*127.46*) | 137.80 (*123.92*) |
| *train_ALL* | 78.93 (*24.14*) | 77.43 (*28.53*) | 79.71 (*22.45*) | 79.65 (*21.45*) | 45.21 (*88.18*) | 66.76 (*111.23*) | 35.25 (*76.39*) | 33.62 (*76.92*) |
| *train_GLI* | 80.04 (*20.89*) | 74.68 (*25.04*) | 82.18 (*21.10*) | 83.27 (*16.54*) | 39.75 (*74.17*) | 77.13 (*94.61*) | 23.07 (*63.99*) | 19.05 (*63.92*) |
| *train_ftSSA* | 82.18 (*21.68*) | 90.81 (*12.59*) | 79.58 (*24.78*) | 76.14 (*27.67*) | 38.65 (*76.77*) | 14.66 (*47.99*) | 42.26 (*82.57*) | 59.04 (*99.73*) |

[a]Metrics reported as mean (*standard deviation*) among different cases of the validation set.

As expected, training a model with only 60 samples (*train_SSA*) is not sufficient, and most validation segmentations were obtained at a chance level (50%). To provide performance estimates on similar data as well as a comparison for other training set compositions, the low-resolution naive *train_GLI* model was submitted for validation in both Task 1 (BraTS-Adult Glioma) and Task 2 (BraTS-Africa). Despite being trained only for 100 epochs on one-fold, this model performed relatively well on the Task 1 validation dataset comprised of 219 glioma patients in terms of performance metrics (mean DSC of 79.81, 82.12, 78.48 and HD95 of 54.04, 28.80, 42.90 for WT, TC and ET, respectively). Average performance metrics across tumour sub-regions achieved on Task 2 validation data were good. As the intention was to use this as a pre-trained base to further train with SSA data, and inspection of internal validation loss indicated it was stabilising, we deemed the model sufficient for subsequent fine-tuning with BraTS-Africa data. Fine-tuning of the trained BraTS-Adult Glioma data with the

BraTS-Africa data across a 5-fold training procedure resulted in a similar mean Dice score across sub-regions.

Further investigation of sub-region performance showed that despite the same mean DSC, the *train_GLI* and *train_ftSSA* models had varied performance for different sub-regions. The *train_GLI* model struggled with delineating the WT from non-tumour tissue, with 40% of cases achieving below chance DSCs. Conversely, the *train_ftSSA* model appeared to do well in this region, but struggled with the ET regions, with 26% of subjects scoring below chance. The ET sub-region is traditionally the most difficult to precisely segment given its size and overlap with surrounding sub-regions. The HD95 also provides a clearer indication of a model's ability to delineate sub-regions, as it better estimates performance on small or low-quality segmentations and when contours are important. Looking at differences in these scores, we see that the *train_ftSSA* model on average was able to predict the whole tumour boundary much better (HD95 ~63 units lower) than the *train_GLI* model. Conversely, the *train_GLI* model is, on average, better able to delineate ET sub-region than *train_ftSSA* does (HD95 ~40 units lower). Figure 3 shows two cases where these differences are clearly seen.

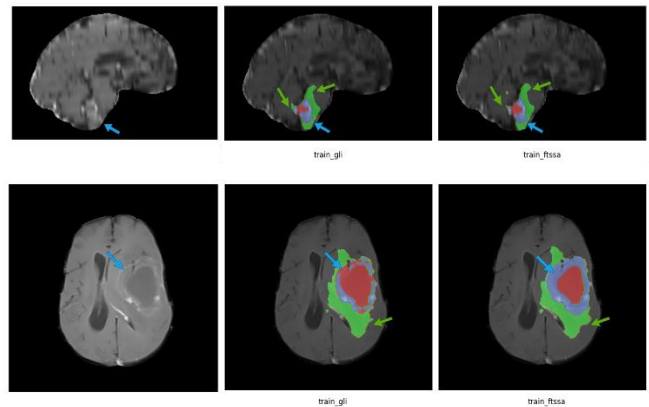

**Fig. 3.** T1Gd of two cases highlighting segmentation differences in ED (green) and ET (blue).

An outlier case was identified where all models struggled to accurately segment the TC sub-region (mean DSC ranging 22.33-49.04) likely due to the confounding presence of ET and TC like voxel intensities within the oedematous tissue in the posterior region of the cortex. This case heavily influenced the *train_ftSSA* scores more than the other models: removing this outlier resulted in DSC increases by 4.13 and 3.43 (with a corresponding drop in HD95 by 13.04 and 8.04) for ET and TC sub-regions, respectively. Furthermore, this outlier also demonstrated improvements in sub-region performance of both ET and TC for the *train_SSA* and *train_ALL* models, albeit weaker than with *train_ftSSA*. Table 4 shows a full comparison of differences in average DSC and HD95 scores. Only performance in the TC sub-region was impacted in the *train_GLI* model. Additionally, one case consistently achieved extremely poor segmentation results for all models with mean DSC across sub-regions ranging 35.40-37.02 and HD95 229.37-229.89, as shown in Figure 4.

**Table 4.** Differences in Dice (DSC) and Hausdorff95 (HD95) after excluding outlier case.

| Model | DSC | | | | HD95 | | | |
|---|---|---|---|---|---|---|---|---|
| | Ave | WT | TC | ET | Ave | WT | TC | ET |
| *train_SSA* | 1.01 | -2.43 | 2.62 | 2.84 | -1.51 | 8.82 | -5.94 | -7.41 |
| *train_ALL* | 1.05 | -1.26 | 2.19 | 2.23 | 1.10 | 4.57 | -0.66 | -0.61 |
| *train_GLI* | **0.25** | -1.43 | **2.38** | -0.20 | **1.67** | 5.31 | **-1.50** | 1.21 |
| *train_ftSSA* | **2.40** | -0.35 | **3.43** | **4.13** | **-11.61** | 0.87 | **-8.04** | **-13.04** |

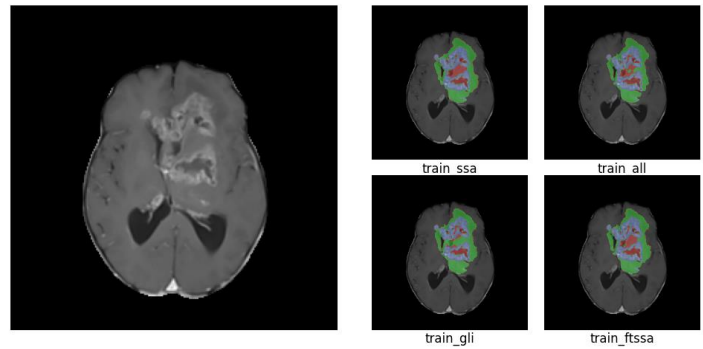

**Fig. 4.** T1Gd of outlier case where poor performance was seen across models for all labels.

Figure 5 depicts a subject for which extremely good scores were obtained on the ET and TC segmentations (mean DSC >89 and HD95<2.00) but not for the WT region. Traditionally, it is more difficult to precisely segment the ET, which is usually a small region. The whole tumour itself should be more easily identifiable. Mean DSC (HD95) achieved was 72.22 (9.273), 0.125 (285.335), 41.33 (190.082) and 89.49 (1.414) for each model depicted in panel 1 from left to right, respectively. Visually we see that most models primarily struggled with delineating oedematous tissue (ED), of which there is very little.

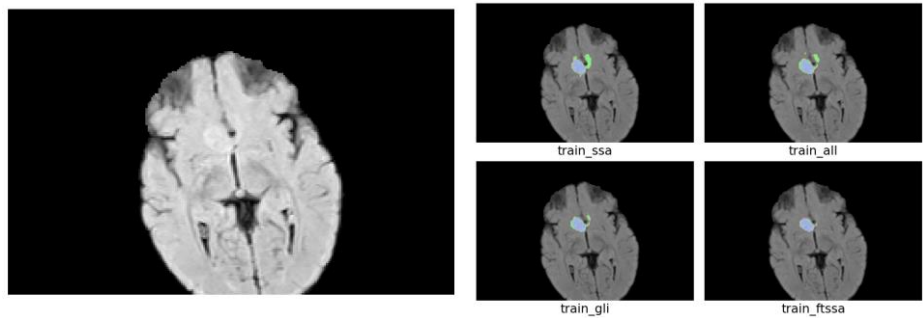

**Fig. 5.** T2-FLAIR showing varied ED (green) segmentation across all models.

In reviewing the performance of all models, we also considered that the whole tumour comprises oedematous tissue, which is harder to differentiate in low-resolution scans. It is likely that without being exposed to the low-quality data from SSA, the *train_GLI* model struggled to precisely identify the boundaries of tumours with extensive oedema. We therefore submitted the model pre-trained with the larger BraTS-Adult Glioma dataset and further trained on the smaller SSA dataset (*train_ftSSA*)[5]. This model performed well on an unseen external test dataset comprised of 20 patients with scans from SSA, ranking 2nd globally. Final DSC and HD95 scores achieved in the testing phase are not publicly available at the time of writing.

## 4   Discussion

Taken together, the results from our experiments emphasise that current state-of-the-art models cannot be implemented directly to SSA data, as the limited training data available will result in overfitting the model. Our final model performed extremely well with a single pre-training fold of 100 epochs and fine-tuning with a 5-fold cross-validation. However, Futrega et al. [37] demonstrated that a 10-fold cross validation allows for more accurate segmentation performance. We only validated 5 folds for *train_SSA* and *train_ftSSA* and did not run cross-validation on the *train_ALL* and *train_GLI* models. It is therefore likely that more extensive cross-validation could yield a more accurate comparison between each model. Yet, the high-ranking performance of our final model reiterate what has been previously demonstrated in simulated cases [39]: state-of-the-art models can be improved with a wider range of data through federated learning, providing opportunity for institutes in low-resource setting to re-train on a smaller set of local data without the need for obtaining data from external sources. However, these results need to be interpreted with caution, as they may be confounded by the limited sample size of the training and validation datasets. Furthermore, going forward, future works should involve both integrating more extensive datasets sourced from various regions within Africa and implementing stronger data augmentations related to scanner artefacts when training on high quality MRI scans. These steps are vital to the development of a model capable of generalising across image qualities seen in varied clinical settings.

**Acknowledgements.** This work was supported by the Lacuna Fund for Health and Equity (PI: Udunna Anazodo, 0508-S-001) and the National Science and Engineering Research Council of Canada (NSERC) Discovery Launch Supplement (PI: Udunna Anazodo, DGECR-2022-00136). The authors gratefully acknowledge the computational infrastructure support provided by the Digital Research Alliance of Canada (The Alliance) and the knowledge translation support through the McGill University Doctoral Internship program. The authors would like to express their appreciation to the global instructors who contributed to the Sprint AI Training for African Medical Imaging Knowledge Translation (SPARK) Academy and Africa-BraTS BrainHack2023 training initiatives. Special thanks go to Linshan Liu for administrative support to the SPARK

---

[5] The code and docker file used for final submission can be found at: https://github.com/CAMERA-MRI/SPARK2023/tree/main/SPARK_UNN

Academy training and capacity building activities, from which the authors benefited. Finally, the authors extend their gratitude to the McMedHacks team for their continuous support throughout the program, including the provision of foundational programming material and their efforts in enabling in-person training.

**Author Contributions.** ARA & AS: *Conceptualisation, Methodology, Software, Validation, Formal Analysis, Investigation, Writing, Visualisation;* PJ: *Writing - Review & Editing, Visualisation, Statistics;* CR, DZ, & UA: *Funding Acquisition, Project Administration, Resources, Data Curation, Conceptualisation, Writing - Review & Editing;* DM & TEMM: *Resources, Writing - Review & editing;* SQ: *Supervision, Conceptualisation, Software, Writing - Review & Editing*